\documentclass{cas-dc}

\usepackage[numbers]{natbib}

\usepackage{natbib}

\usepackage{multirow}
\usepackage{hyperref}
\hypersetup{colorlinks=true,citecolor=blue,linkcolor=blue}
\usepackage{tikz}
\usepackage{float}

\usepackage[justification=centering]{caption}

\newcommand*{\circled}[1]{\lower.7ex\hbox{\tikz\draw (0pt, 0pt)%
		circle (.4em) node {\makebox[1em][c]{\small #1}};}}

\def\tsc#1{\csdef{#1}{\textsc{\lowercase{#1}}\xspace}}
\tsc{WGM}
\tsc{QE}
\tsc{EP}
\tsc{PMS}
\tsc{BEC}
\tsc{DE}

\begin{document}
	\let\WriteBookmarks\relax
	\def\floatpagepagefraction{1}
	\def\textpagefraction{.001}
	\shorttitle{HIE for the link prediction}
	\shortauthors{Jin Liu et~al.}
	
	\title [mode = title]{Joint embedding in Hierarchical distance and semantic representation learning for link prediction }   
	
	
	

	
	\author{Jin Liu}[style=chinese]
	
	\author{Fengyu Zhou}[style=chinese]
	\cormark[1]
	\ead{zhoufengyu@sdu.org}
	\author{Jianye Chen}[style=chinese]
	\author{Chongfeng Fan}[style=chinese]
	
	\address{School of control science and Engineering, Shandong university, Jinan 250011, China}
	
	
	%
	
	\cortext[cor1]{Corresponding author}

	
	
	\begin{abstract}
		The link prediction task aims to predict missing links in the knowledge graph and is essential for the downstream application.  
		Existing well-known models deal with this task by mainly focusing on representing knowledge graph triplets in the distance space or semantic space. However, they can not fully capture the information of head and tail entities, nor even make good use of hierarchical level information.
		Thus, 
		in this paper, we propose a novel knowledge graph embedding model for the link prediction task, namely, HIE, which models each triplet (h, r, t) into distance measurement space and semantic measurement space simultaneously. Moreover, to leverage the hierarchical information of entities and relations, HIE is introduced into hierarchical-aware space for better representation learning.  
		Specifically, we apply distance transformation operation on the head entity in distance space to obtain the tail entity instead of translation-based or rotation-based approaches.
		Experimental results of HIE on four real-world datasets show that HIE outperforms several existing state-of-the-art knowledge graph embedding methods on link prediction task and deals with complex relations accurately.      
	\end{abstract}
	%
	
	
	
	\maketitle
	
	\section{Introduction}
\begin{figure}
		\centering
		\includegraphics[scale=0.45]{ 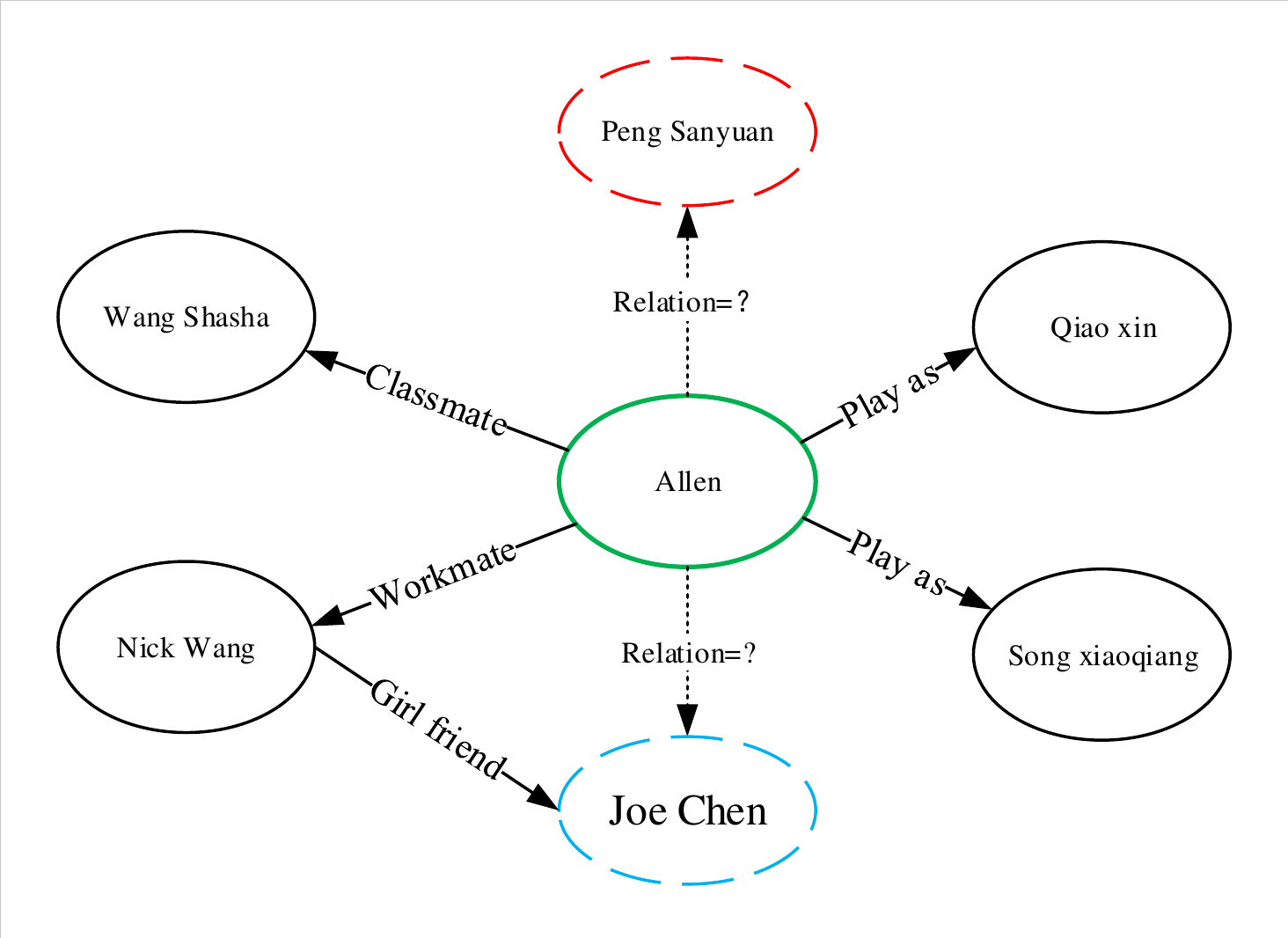}
		\caption{Illustration of a small KG for Allen, where the relation between Joe Chen and Allen, Peng Sanyuan and Allen are needed to infer via KG embedding methods.}
		\label{fig:zhishi}
	\end{figure}

	Knowledge graphs (KGs) have been an essential component in artificial intelligence (AI) and applied to many downstream applications, such as question answering\cite{xiong2021knowledge}\cite{damchen}, sentiment analysis\cite{ZHAO2021107220} and image caption\cite{imagecap}. Numerous well-known large-scale knowledge graphs, such as WordNet\cite{miller1995wordnet}, Freebase\cite{bollacker2008freebase} and YAGO\cite{suchanek2007yago}, composed of structural information from manifold human knowledge, stored in a directed graph, have been widely employed for various domain in the past decades. The form of knowledge fact is typically triplet, i.e. (head entity, relation, tail entity) or (subject entity, relation, object entity), or (\textit{h}, \textit{r}, \textit{t}) in short.  Specifically, the head or tail entity denotes the concepts or definitions, and the relation represents the relationship between entities. Taking a triplet (Allen, Workmate, Wang kai) from Fig.\ref{fig:zhishi} as an example, it describes the fact that Wang kai is the workmate of Allen.
	
	Although previous KGs, such as Wordnet, Freebase and YAGO are composed of massive facts, they inevitably suffer from the incompleteness of the existing knowledge graphs. As the KG shown in Fig.\ref{fig:zhishi}, we can get the corresponding facts of the triplets that are circled in black and connected to Allen. However, conditioned on the KG, we do not know what relationship between Allen and Joe Chen, nor the relationships among those are connect to Allen. Thus, accurately predicting the missing links between known entities, i.e., link prediction task, has gained much more attention. Many knowledge graph embedding (KGE) methods are proposed to learn the continuous low-dimensional embedding expressions of entities and relations to fulfill this task. According to the form of score functions, previous KGE methods can be roughly classified into three categories: (1) Translation-based models (2) Semantic matching models (3) Deep neural network-based models. These approaches achieve promising results to some extent. However, since these methods can only capture structural information or semantic information, they lack the ability to extract hierarchical features and take advantage of two kinds of information, which potentially limits their performance on the link prediction task.
	
	
	As the seminal work in KGEs, TransE\cite{transE} projects knowledge graphs into low-dimensional space efficiently and obtains satisfactory results in 1-to-1 tasks. The modeling ability of TransE may be because it could capture a lot of structural information in the distance measurement space. However, TransE fails to model other complex relation properties of 1-to-N, N-to-1 and N-to-N. To solve such a problem, TransR\cite{transR} utilizes the relationship-specific mapping matrix to project head and tail entities into relation-specific space. Nevertheless, there may be a little differences between tail entities and head entities under the same relationship. Besides, TransR ignores the structural information of the distance measurement space and the latent semantic information between entity-relations, which can also impair the expression ability of the model. Another exciting work M-DCN to deal with such a problem was proposed in \cite{mdcn}. The model uses convolutional kernels and embedding combination to capture triplet-level semantic information. However, the structural features are weakened, and the way of entity combination in M-DCN can omit some original embedding information.             
	
\begin{figure}
		\centering
		\includegraphics[scale=0.5]{ 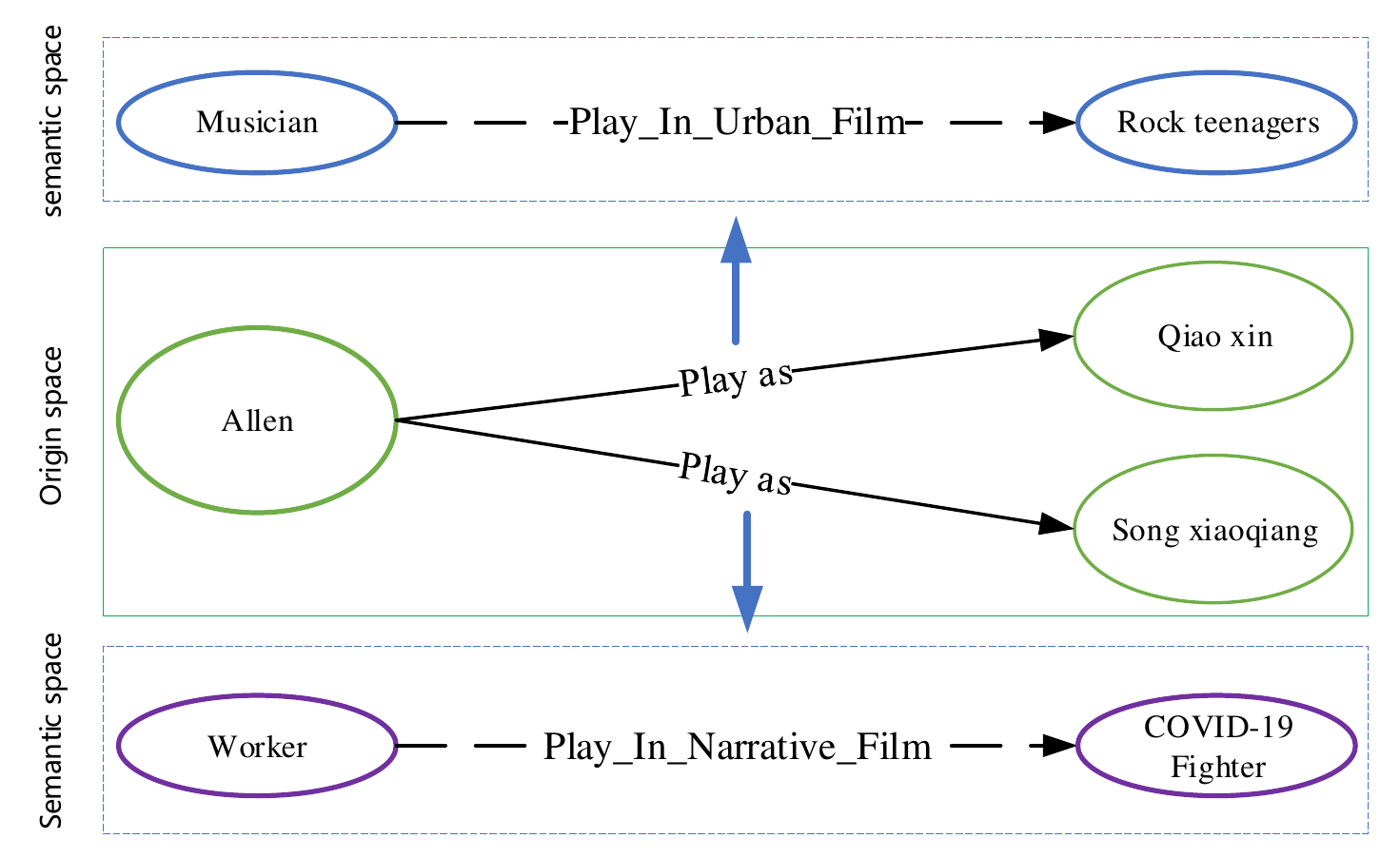}
		\caption{Illustration of the projection from origin space to entity-relation semantic space.}
		\label{fig:jieshi}
	\end{figure}

	Thus, to build an expressive KGE model and deal with the link prediction task, we propose a novel knowledge graph embedding model named HIE. Unlike the previous works, we project each entry (h, r, t) into the distance measurement space and semantic measurement space to keep structural and entity-relation's semantic information through the corresponding mapping matrix. Consequently, the model can leverage more complex features between entities and their relationships. Taking the relationship \textit{play as} in Fig.\ref{fig:zhishi} as an example, in the distance measurement space, models may get Emb(Qiao xin) $\approx$ Emb(Song xiaoqiang). While in the semantic measurement space, the projected tail entities (i.e., Qiao Xin and Song xiaoqiang) may be close via the play\_as-specific matrix by using TransR or M-DCN and the relation may not be suitable for the projected entities. Thus, the modeling ability of methods may be limited by only projecting the entities and relations in distance measurement or semantic space. Comparatively, HIE projects the head entity, relation and tail entity into fine-grained semantic space, which is shown in Fig.\ref{fig:jieshi}. That's to say, we may get the two plausible triplets (Allen(musician), play\_in\_urban\_film, Qiaoxin(rock teenagers)) and (Allen(worker), play\_in\_narrative\_film, Song xiaoqiang(covid-19 fighter)). As for the relationship \textit{Classmate}, the projected embedding in the distance measurement space keeps the structural information of the original triplet. Therefore, HIE can distinguish the triplets easily under the different relations. In addition, motivated by the TKRL model\cite{xie2016representation}, which models entities with hierarchical types. We jointly extract hierarchical geometric information (from distance measurement) and semantic information (from semantic measurement) to capture more triplet features, which can also be treated as a multi-view way to use knowledge to solve the link prediction task. In the empirical experiments, we compare HIE with several state-of-the-art KGE methods, including translation-based, neural network-based and semantic matching-based models. The results on the benchmark datasets show that our proposed HIE outperforms the baseline model in the link prediction task and outperforms several complex relations modeling task.       
	
	In conclusion, our main contributions are summarized as follows:
	\begin{itemize}
		\item We propose a novel KGE model, namely, HIE, which keeps structural geometric information and captures semantic information by projecting each triplet (h, r, t) into both distance measurement and semantic measurement space to learn the knowledge representation jointly.
		
		\item We use distance transformation operation instead of translation-based or rotation-based projection to obtain tail entity. And adopt the semantic extractor to learn fine-grained entity and relation representations in different semantic spaces. Moreover, we extract hierarchical geometric information and semantic information for the representation learning process to capture more information without extra data.
		
		\item To the best of our knowledge, we are the first to combine distance measurement with semantic measurement and further employ the hierarchical information for the model to address the link prediction task.
		
		\item We conduct our experiments extensively on four representative benchmark datasets WN18\cite{transE}, WN18RR\cite{conve}, FB15k-237\cite{convkb} and YAGO3-10\cite{RotatE}. The evaluation results illustrate that HIE outperforms the compared several state-of-the-art methods on the link prediction task. More importantly, HIE shows its robustness for empirical results via comprehensive ablation studies.     
	\end{itemize}
	
	The rest of this article is organized as follows. Related work is briefly reviewed in Sec.\ref{relatedwork}. In Sec.\ref{ourmodel}, the proposed HIE is presented in detail. In Sec.\ref{myexp}, extensive experiments are conducted on four real-world datasets. Also, the experimental protocols, evaluation metrics and the results on link prediction and complex relations are reported. The conclusion and future work are discussed in Sec.\ref{myconclusion}. 
	
	\section{Related work}\label{relatedwork}
	Knowledge graph embedding has been widely studied as one of the most critical parts for solving the link prediction task in recent years. According to the score functions and representation approaches, existing KGE models can be roughly divided into three categories: Translation-based models, semantic matching models, deep neural network models. Table.\ref{tbl1} reports several state-of-the-art KGE models and their parameters. 
	
	\begin{table*}[!htbp] 
		\centering
		\caption{Summary of several KGE models' score function: where $\bar{h}$ denotes the conjugate operation over $h$, $<h,t>$ denotes the Hermitian dot product and equals $\bar{\text{h}}^{T}t$. $\circ$ represents the Hadamard product. f$_{D}$ denotes the  score function of discriminator. $\textit{p}$(\textbf{h},\textbf{r};\textbf{c})=(\textbf{c}$\circ$\textbf{h})$\star$\textbf{r}, where $\star$ denotes the circular correlation, c is a fixed vector. $\ast$ denotes convolution operation, $\oplus$ represents the concatenate operation, $\circledast$ denotes the depthwise circular convolution operation.}\label{tbl1}
		\scalebox{0.9}{
			\begin{tabular}{cccc}   
				\toprule  
				Category & Model name & Score function & Parameters\\
				\midrule  
				\multirow{6}{*}{Translation-based models} &TransE &-$\lvert \lvert$\textbf{h} + \textbf{r} - \textbf{t}$\rvert \rvert_{L_1/L_2}$  &\textbf{ h,r,t}$\in$ $\mathbb{R}^{d}$  \\
				&TransD &-$\lvert \lvert$(\textbf{r$_{p}$}\textbf{h$_{p}^{T}+\mathrm{I}$})\textbf{h}+\textbf{r}-(\textbf{r$_{p}$}\textbf{t$_{p}^{T}+\mathrm{I}$})t$\rvert \rvert_{2}^{2}$ & \textbf{h},\textbf{t},\textbf{w$_h$},\textbf{w$_t$}$\in$ $\mathbb{R}^{d}$,\textbf{r},w$_r$$\in$ $\mathbb{R}^{k}$ \\ 
				&TransR &\textbf{-$\lvert \lvert$M$_r$h+r-M$_r$t$\rvert \rvert_{2}^{2}$} &\textbf{h,t,r}$\in$ $\mathbb{R}^{d}$,M$_r$ $\in$ $\mathbb{R}^{d\times d}$\\
				&ComplEx &Re(\textbf{<r, h,  $\bar{\text{t}}$>}) &\textbf{h,r,t}$\in$ $\mathbb{C}^{d}$ \\
				&RotatE &\textbf{$\lvert \lvert$h $\circ$ r - t $\rvert \rvert$ }&\textbf{h,t,r},w$_r$$\in$ $\mathbb{R}^{d}$ \\
				&TorusE &$min_{(x,y)\in ([\text{h}]+[\text{r}]) \times [\text{t}] }$ $\lvert \lvert$x - y$\rvert \rvert_{i}$ &[\textbf{h}],[\textbf{r}],[\textbf{t}]$\in$ $\mathit{T}^{n}$ \\
				\hline
				\multirow{4}{*}{Semantic matching models}&DistMult &\textbf{h$^{T}$}diag(M$_r$)\textbf{t} &\textbf{h,r,t}$\in$ $\mathbb{R}^{d}$ \\
				&RESCAL &\textbf{h$^{T}$}M$_r$\textbf{t} &\textbf{h,t}$\in$ $\mathbb{R}^{d}$,M$_r$$\in$$\mathbb{R}^{d\times d}$ \\
				&HolEx & $\sum_{j=0}^{l}p$(\textbf{h,r;c}$_j$)t &\textbf{h,r,t}$\in$ $\mathbb{R}^{d}$ \\
				&SimplE &$\frac{1}{2}$(\textbf{h}$\circ$\textbf{rt} + \textbf{t}$\circ$\textbf{r}$^{'}$\textbf{t}) &\textbf{h,r,t,r$^{'}$}$\in$ $\mathbb{R}^{d}$ \\
				\hline
				\multirow{7}{*}{Deep neural network models}&KBGAN &\textit{f$_D$}(\textbf{h,r,t}) &\textbf{h,r,t}$\in$ $\mathbb{R}^{d}$ \\
				
				&ConvE &\textit{f}(vec(\textit{f}([$\bar{\text{\textbf{h}}}$;$\bar{\text{\textbf{r}}}$] $\ast$ $\Omega$))\textit{W})\textbf{t} &\textbf{h,r,t}$\in$ $\mathbb{R}^{d}$ \\
				&ConvKB &concat(\textit{f}([\textbf{h,r,t}] $\ast$ $\Omega$))$\textit{W}$ &h,r,t$\in$ $\mathbb{R}^{d}$ \\
				&M-DCN &$\sigma$(\textit{f}(vec(\textit{f}([$\overline{\text{\textbf{h}}\oplus\text{\textbf{r}}}$]$\ast$\textbf{w$_r^i$}))$\textit{W}$)\textbf{t}+b) &\textbf{h,r,t}$\in$ $\mathbb{R}^{d}$, w$_r^i$$\in$$\mathbb{R}^{1\times2}$ \\
				&InteractE &\textit{g}(vec(\textit{f}($\phi$([\textbf{(h,r)}])$\circledast$w))\textit{W})\textbf{h} &\textbf{h,r,t}$\in$ $\mathbb{R}^{d}$, \textbf{w}$\in$$\mathbb{R}^{k \times k}$  \\
				\bottomrule 
			\end{tabular}
		}
	\end{table*}
	
	\subsection{Translation-based models}
	TransE\cite{transE} is regarded as the most representative translation-based model, interpreting relationships as a translation from the head entity to the tail entity. For a triplet (\textit{h},\textit{r},\textit{t}), it attempts that \textbf{h}+\textbf{r}$\approx$\textbf{t} and employs $f_r$(\textbf{h},\textbf{t})= -$\lvert \lvert$\textbf{h} + \textbf{r} - \textbf{t}$\rvert \rvert_{L_1/L_2}$ as the corresponding score function under $L_1$ or $L_2$ constraints. Despite its simplicity and efficiency, TransE couldn't obtain well results on complex relations, such as 1-to-N, N-to-1 and N-N. In order to address such issues, TransH\cite{transH} projects the entities into the relation hyperplane and performs a translation operation on the transformed embeddings. TransR\cite{transR} employs projection spaces for entities and relations via a relation-specific mapping matrix M$_r$ and performs translation into the relation-specific space. To improve the performance of TransR, Ji et al.\cite{transD} proposed a more fine-grained model TransD, in which entities and relations are composed of two basic vectors. One is for capturing the semantic meanings, and the other one for constructing dynamic mapping matrices M$_{rh}$=\textbf{r$_p$h$_{p}^{T}$}+I$^{m\times n}$ and M$_{rt}$=\textbf{r$_p$t$_{p}^{T}$}+I$^{m\times n}$ via the projection vectors\textbf{ h$_p$, t$_p$} and \textbf{r$_p$}. 
	
	Recently, there have been some variants of the translation-based model by representing the head entities and relations into the complex vector space and manifold space.  ComplEx \cite{compleX} firstly applies complex value embedding approach to entities and relations, which attempts to model symmetric and antisymmetric relations via Hermitian dot product between real vectors and its conjugate. However, it can't deal with the composition relation pattern. To solve this problem, RotatE\cite{RotatE} models the relations as an element-wise rotation rather than a translation from head entities to tail entities in complex space, and the corresponding score function is defined as $\lvert \lvert$\textbf{h} $\circ$ \textbf{r} - \textbf{t} $\rvert \rvert$. Moreover, since TransE exists the problems of regularization and an unchangeable negative sampling ratio, TorusE\cite{torusE} projects the entities and relations on a compact Lie group in torus space, and the score function is defined as $min_{(x,y)\in ([\text{h}]+[\text{r}]) \times [\text{t}] }$ $\lvert \lvert$x - y$\rvert \rvert_{i}$ with the projection: $\mathbb{R}^{n} \to \mathit{T}^{n}$, $x \mapsto [x] $, where [\textbf{h}], [\textbf{r}], [\textbf{t}] $\in$ $\mathit{T}^{n}$. Additionally, TorusE adopts the similar relation translation, i.e., [\textbf{h}] + [\textbf{r}] $\approx$ [\textbf{t}]. 
	%
	However, the above models simply project both entities and relations into the same embedding space. Thus, they couldn't model the hierarchical characteristics for entities and relations, which leads to weak improvements for link prediction task.
	
	\subsection{Semantic matching models}
	Another way to obtain the embeddings is to match the latent semantic similarity between entities and relations in vector space. SE\cite{semodel}, which is regarded as a linear model, projects the specific relation as a square matrix pair M=(M$_1$, M$_2$) and uses the similarity function $f_r$(\textbf{h,t})=-$\lvert \lvert$M$_1$\textbf{h} - M$_2$\textbf{t}$\rvert \rvert_{L_1/L_2}$ to determine a plausible triplet. As for the bilinear model, RESCAL\cite{rescal} is a representative tensor factorization model to capture the latent structure information, where each relation is factored as a dense square matrix M$_r$ to associated with the entity vector embedding. However, due to the three-way rank-r factorization, it has quadratic growth parameters and tends to overfit\cite{rescalover}. DistMult\cite{DistmulT}, which extends RESCAL by modifying the bilinear function, proposes a simple embedding approach through a bilinear product and replaces the complex relation matrix with an easy-to-train diagonal matrix. However, when dealing with the symmetric relations, i.e., (h,r,t) and (t,r,h), DistMult exhibits the same results for triplets and lacks adequate expressive power. By employing circular correlation of entity and relation embeddings, HolE\cite{hole} can capture rich embedding interaction information and maintain the computing efficiency as DistMult. To further improve the representation ability of HolE, Xue et al.\cite{holeX} proposes HolEx that interpolates the full product matrices to achieve a lower dimensionality. SimplE\cite{simplE} introduces a simple tensor factorization approach by calculating the average score of the inverse triplets to be full-expressive, i.e., (\textbf{h, r, t}) and (\textbf{t, r$^{-1}$, h}), where\textbf{ r$^{'}$} represents the inverse relation embeddings. 
	
	Although these semantic models can capture semantic information between entities and relations, they are prone to suffering overfitting due to the model redundancy in low dimension embedding space. Further, the structural information of the triplets can not be fully utilized.       
	
	\subsection{Deep neural network-based models}
	Recently, deep neural networks have been applied in various domains successfully and attracted more and more attention in solving link prediction task. Representative neural tensor work (NTN)\cite{ntn} takes entity embeddings and relation vectors as input and measures the plausibility of a triplet associated with a particular relationship by combining multi-layer perceptrons (MLPs) with bilinear models. ConvE\cite{conve} and ConvKB\cite{convkb} adopt a convolution neural network to model interactions between entities and relations for the link prediction task. The difference between them is that ConvE uses 2D convolution on a reshaped entity and relation embedding matrix while ConvKB encodes the concatenation of entity and relation embeddings without reshaping. Apart from CNN, Nguyen et al.\cite{capsE} proposes  CapsE, which takes each triplet as a 3-column matrix and adopts Capsule Net to capture the information of embeddings \cite{capsNetwork}. However, these models suffer from three problems: (1) it is insufficient in dealing with complex relations; (2) the number of interactions that they can capture is limited; (3) the structural information between entities and relations can be ignored easily. Thus, Zhang et al.\cite{mdcn} presents a multi-scale dynamic convolutional network to explore the entity and relation embedding characteristics for the first problem, namely, M-DCN. As for the second problem, InteractE\cite{interacte} increases the number of additional interactions via feature permutation, a novel feature reshaping, and circular convolution. For the purpose of modeling various relational patterns, capturing structural information and improving the computation efficiency of deep neural networks, Liu et al.\cite{aprile} proposes AprilE by introducing a triple-level self-attention approach. Graph convolution networks (GCNs) have been used as another way to capture the graph-structured and embedding information simultaneously. R-GCN\cite{rgcN} uses relation-specific GCN to model the directed knowledge graph. KBGAN\cite{cai-wang-2018-kbgan} combines KGE with generative adversarial networks (GANs) to model the entities and relations, and the score function is defined in Table.\ref{tbl1}. 
	
	Although neural network-based models can capture semantic information of triplet more easily than translation-based and semantic matching models, they still suffer from large model parameters, and the inability to make good use of the global structure.
	
	\subsection{models with hierarchical structural information}
	To make use of auxiliary information from external knowledge for the link prediction task, several models have introduced hierarchical structures in recent years. TKRL projects entities by type information matrices, which are composed of weighted and recursive hierarchical type encoders\cite{xie2016representation}. HA-RotatE projects entities with different moduli in the different hierarchical levels \cite{wang2021hierarchical}. MuRP takes full advantage of hyperbolic embeddings to represent hierarchical relational graph data according to the relation-specific parameters \cite{balazevic2019multi}. Comparatively, HAKE learns the hierarchical entity embeddings in the polar coordinate system.
	
	The above hierarchical-based learning methods only capture entity or relation features in their embedding space. However, they fail to keep the geometric information or semantic information simultaneously. Thus, they can only show advantages in dealing with complex relation properties or relation patterns.   
	
	\section{Our Model}\label{ourmodel}
	\begin{table}[!htbp] 
		\centering
		\caption{Notations and Explanations of HIE which are used in this paper.}\label{tblnotion}
		\scalebox{0.9}{
			\begin{tabular}{cc}   
				\toprule  
				Notations & Short explanations\\
				\midrule  
				$\mathcal{G}$ & Knowledge graph \\
				$\mathcal{E},\mathcal{R},$T & entities, relations and triplets   \\
				\textbf{h,r,t} & head entity, relationship and tail entity \\
				$f_r$(\textbf{h,t}) & score function \\
				$\lambda_{1}, \lambda_{2}$, $\lambda_{\cdots}$ & the weights of different hierarchical level\\
				$\circ$ & Hadamard product \\
				$\lvert\lvert \cdot \rvert\rvert_{L_1/L_2}$ & $L_{1}$ or $L_{2}$ norm \\
				$diag$($\cdot$)  & the diagonal projection matrix \\
				\bottomrule 
			\end{tabular}
		}
	\end{table}  
	This section presents our proposed model HIE for the link prediction task, which can capture the position and semantic information in hierarchical level representation spaces. Before introducing the HIE model, the basic mathematical notions of model are summarized in Table.\ref{tblnotion}. 
	
	\subsection{Problem formulation}
	A knowledge graph consists of triplets, which can be expressed as $\mathcal{G}$= ($\mathcal{E},\mathcal{R},$T), where $\text{\textit{h}}_{0,\cdots,|\mathcal{E}|}$ ,$\text{\textit{t}}_{0,\cdots,|\mathcal{E}|}$ $\in$ $\mathcal{E}$ and $\text{\textit{r}}_{0,\cdots,|\mathcal{R}|}$ $\in$ $\mathcal{R}$. $|\mathcal{E}|$ and $|\mathcal{R}|$ denote the number of entities and relations, respectively. For a link prediction task, the model aims to predict the head entity with a given tail entity (?,\textit{r},\textit{t}) or the tail entity with a given head entity (\textit{h},\textit{r},?) to get a valid triplet (\textit{h,r,t}). Moreover, the model needs to learn the low-dimension embeddings for entities and relations.
	
	\subsection{HIE model}
	\subsubsection{Basic structure space}
\begin{figure}[h]
		\centering
		\includegraphics[scale=0.7]{ 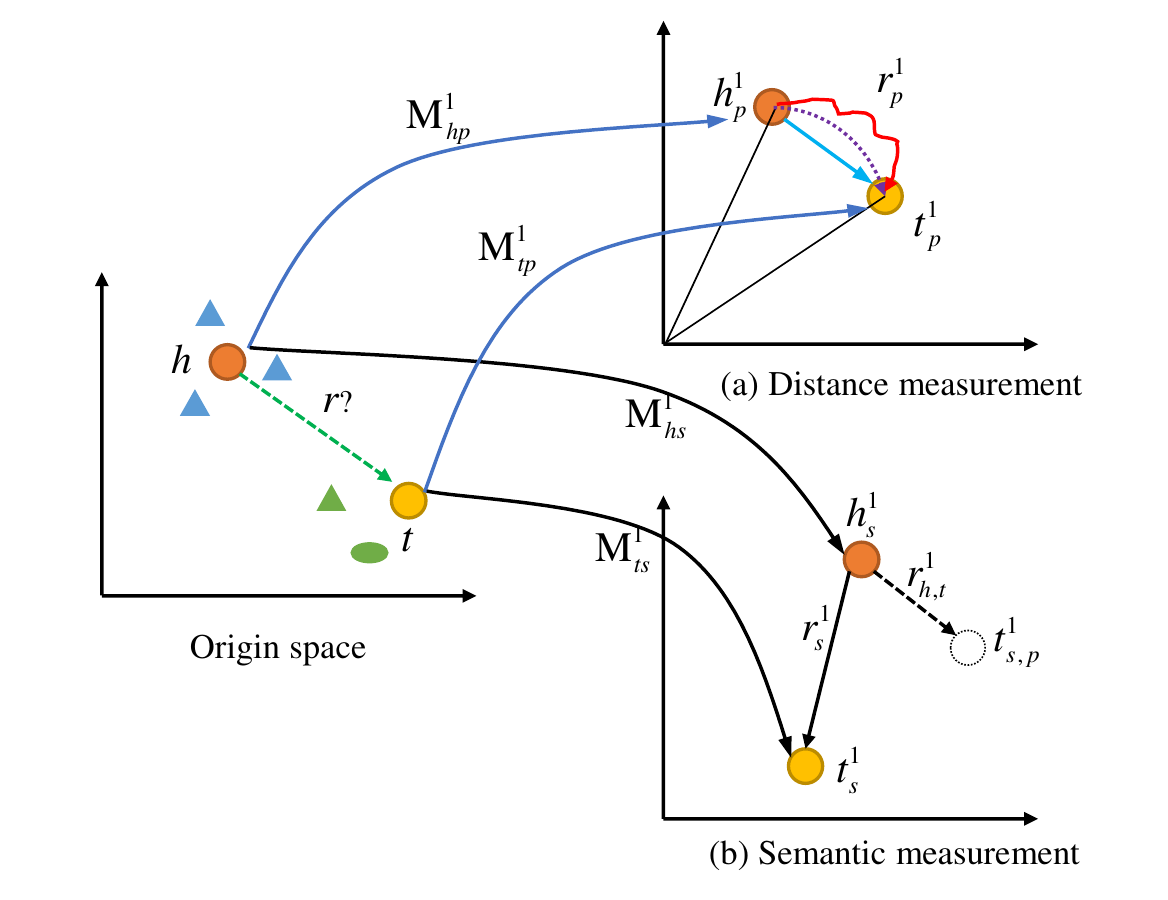}
		\caption{The basic structure of HIE.}
		\label{fig:2}
	\end{figure}

	Unlike the TKRL, which captures hierarchical entity type information, and TranR, which projects entities into relation-specific space, comparatively, we model the triplet into distance measurement and semantic measurement space. The basic distance measurement space and semantic measurement space structure are shown in Fig.\ref{fig:2}. For each entity or relation, we consider that it carries two parts of information. One is the geometric information for the distance measurement of entities in distance space under the specific relation. The other one is used to identify the sub-semantic restrictions in semantic space. It's worth noting that the concept of distance measurement is a general way to determine the position change degree in the embedding space. Take a triplet from Fig.\ref{fig:zhishi}, (Allen, classmate, Wangshasha), the "classmate" is 1-1 relationship in this sub-graph. Thus, entities and relationship embedding in distance measurement space can easily distinguish the triplet, the effect of the semantic is slightly low. While we choose the 1-N relation "play as", the sub-semantic plays a more important role than distance measurement space. Also, the distance measurement space keeps the geometric for the origin triplet as auxiliary information, which can be useful for determining a plausible long path triplet in dealing with complex relation properties\cite{zhang2019quaternion}.
	
	The steps of the proposed method in basic distance measurement space and semantic measurement space are described as follows.
	
	{Firstly}, the entities and relations are randomly initialized as $k$-dimension representation. Then, all of them are equally segmented into two parts. For example head entity h is divided into \textbf{$\text{h}_{0}$} and \textbf{$\text{h}_{1}$}. Here, we consider that \textbf{$\text{h}_{0}$} consists of geometric information and \textbf{$\text{h}_{1}$} consists of semantic information. 
	
	{Secondly}, we project the \textbf{$\text{h}_{0}$} and \textbf{$\text{h}_{1}$} in distance measurement space and semantic measurement space, respectively.
	
	\textbf{Distance measurement space} In this geometric space, we utilize the corresponding matrices to obtain \textbf{$\text{h}_{p}^{1}$}, \textbf{$\text{t}_{p}^{1}$} and \textbf{$\text{r}_{p}^{1}$}, respectively. The element-wise projection in the distance measurement space is defined as:
	\begin{equation}\label{element_point}
		\centering	\left\{ \ \ 
		\begin{split}
			\text{\textbf{h}}_p^1&=diag(\text{M}_{hp}^1) \circ \text{\textbf{h}}_0   \\
			\text{\textbf{t}}_p^1&=diag(\text{M}_{tp}^1) \circ \text{\textbf{t}}_0     \\
			\text{\textbf{r}}_p^1&=diag(\text{M}_{rp}^1) \circ \text{\textbf{r}}_0    
		\end{split}
		\right.
	\end{equation}
	where \textbf{h$_0$}, \textbf{t$_0$} and \textbf{r$_0$} denote the first segment representation of \textit{h}, \textit{r}, and \textit{t}, respectively. $diag(\text{M}_{hp}^s)$ is a diagonal matrix that projects h$_0$ into geometric distance measurement space, $diag(\text{M}_{tp}^s)$ and $diag(\text{M}_{rp}^s)$ denote the same operation on \textbf{t$_0$} and\textbf{ r$_0$}. 
	
	Instead of using rotation or translation operations for knowledge graph representation, a flexible transformation matrix $\text{M}_r^s$ is used to capture the geometric features among entities and relations. As is shown in Fig.\ref{fig:2}, the light blue straight line models the relation as a translation, and the purple curve dotted line treats the relation as a rotation. However, these modeling approaches are not flexible, nor capture more transformed information between entities and relations for dealing with complex relations. Thus, a soft transform matrix is proposed to overcome the problems, which is shown as the red curve solid line. In this way, HIE stores all the possible form of transformation and movement operation parameters through the matrix $\text{M}_r^1$. Moreover, this approach can ensure that the projecting procedure becomes more flexible.
	\begin{equation}\label{mappm}
		\centering
		\text{M}_r^1=\text{M}_0\text{\textbf{r}}_p^1
	\end{equation}
	where $\text{M}_0$ is used to form transformation operation matrix and the dimension of $\text{M}_0$ is $\frac{k}{2}$*1.
	
	Hence, the position distance function in the first level of HIE is shown in Eq.\ref{shallow_dis}.
	
	\begin{equation}\label{shallow_dis}
		\centering
		\text{d}_r^{p1}=\lvert\lvert \text{\textbf{h}}_p^1\text{M}_r^1 -\text{\textbf{t}}_p^1\rvert\rvert_{L_1/L_2}
	\end{equation}
	
	\textbf{Semantic measurement space} In this semantic space, we tend to use sub-semantic to project entities and relations in a fine-grained way. Similarly, the projection in this space is defined as Eq.\ref{element_sem}. 
	
	\begin{equation}\label{element_sem}
		\centering	\left\{ \ \ 
		\begin{split}
			\text{\textbf{h}}_s^1&=diag(\text{M}_{hs}^1) \circ \text{\textbf{h}}_1    \\
			\text{\textbf{t}}_s^1&=diag(\text{M}_{ts}^1) \circ \text{\textbf{t}}_1     \\
			\text{\textbf{r}}_s^1&=diag(\text{M}_{rs}^1) \circ \text{\textbf{r}}_1     
		\end{split}
		\right.
	\end{equation}
	where \textbf{h$_1$, t$_1$,} and \textbf{r$_1$} denote the second segment of h, r, and t, respectively. $diag(\text{M}_{hs}^s)$ is a semantic extractor that projects \textbf{h$_1$} into fine-grained sub-semantic measurement space, $diag(\text{M}_{ts}^s)$ and $diag(\text{M}_{rs}^s)$ denote the same operation on\textbf{ t$_1$} and \textbf{r$_1$}.  
	
	As is illustrated in Fig.\ref{fig:2}(b), \textbf{t$_{s,p}^1$} is the predicted tail entity and \textbf{r$_{h,t}^1$} is the ideal relation embedding. Similar to \cite{type}\cite{atuo}, we also try to have the following equation: \textbf{h$_s^1$} + \textbf{r$_s^1$ }$\approx$ \textbf{t$_s^1$}, i.e. \textbf{r$_s^1$} $\approx$ \textbf{r$_{h,t}^1$}. And \textbf{r$_s^1$} is a projected composition representation in semantic space, and \textbf{r$_s^1$} here denotes a mixed semantic information representation, including types, concepts, and other semantic information. The following Eq.\ref{shallow_sem} is used to obtain the score of triplet.
	\begin{equation}\label{shallow_sem}
		\centering
		\text{d}_r^{s1}=\lvert\lvert \text{\textbf{h}}_{s}^1 + \text{\textbf{r}}_{s}^1 - \text{\textbf{t}}_{s}^1 \rvert\rvert_{L_2}
	\end{equation}
	{Thirdly}, according to the above projections in two spaces, the joint score function in the basic structural level is defined as Eq.\ref{shallow}.
	
	\begin{equation}\label{shallow}
		\centering
		\text{d}_r^1=\alpha \text{d}_r^{p1} +(1-\alpha) \text{d}_r^{s1}
	\end{equation}
	where $\alpha$ is a learnable weight parameter used to take advantage of distance measurement space and semantic measurement space.
	
	\subsubsection{Hierarchical structure representation}
	\begin{figure*}
		\centering
		\includegraphics[scale=1.1]{ 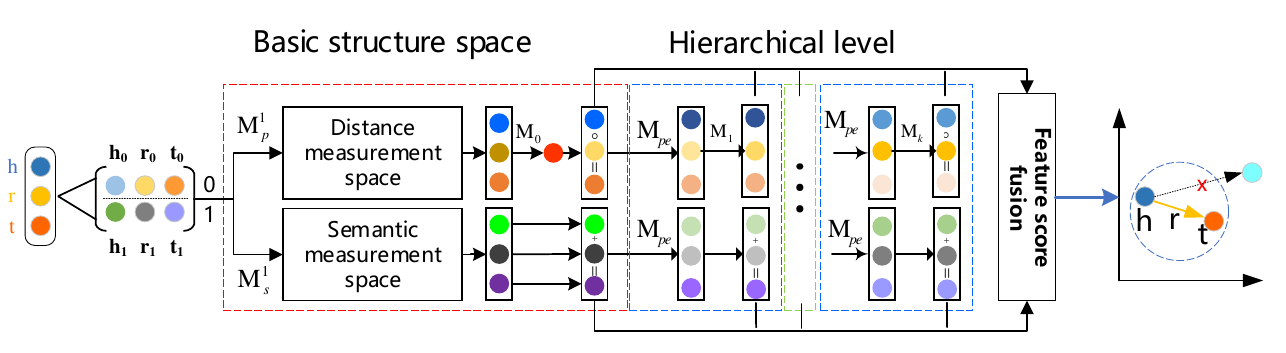}
		\caption{The whole architecture of the proposed HIE.}
		\label{fig:4}
	\end{figure*}
	In order to capture hierarchical information from different space levels, we encoder the basic structure space representation in hierarchical way, which is shown in Fig.\ref{fig:4}. Also, the semantic and position extraction matrix are defined as follows:
	\begin{equation}\label{pos_extract}
		\centering \left\{ \ \
		\begin{split}
			\text{\textbf{h}}_p^2&=\text{\textbf{h}}_p^1\text{M}_{pe} +\textbf{h}_0\\
			\text{\textbf{t}}_p^2&=\text{\textbf{t}}_p^1\text{M}_{pe} +\textbf{t}_0 \\
			\text{\textbf{r}}_p^2&=\text{\textbf{r}}_p^1\text{M}_{pe} +\textbf{r}_0\\
			\text{\textbf{h}}_s^2&=\text{\textbf{h}}_s^1\text{M}_{se} +\textbf{h}_1 \\
			\text{\textbf{t}}_s^2&=\text{\textbf{t}}_s^1\text{M}_{se} +\textbf{t}_1\\
			\text{\textbf{r}}_s^2&=\text{\textbf{r}}_s^1\text{M}_{se} +\textbf{r}_1
		\end{split}
		\right.
	\end{equation}
	where $\text{M}_{pe}$ denotes the geometric extraction matrix that can obtain latent information from the shallow level, $\text{M}_{se}$ is the semantic extraction matrix. The deeper information can be captured via the above matrices.
	
	Also, through similar operations on the new embeddings in the deep level representation space, the position distance function can be defined as:
	
	\begin{equation}\label{deep_dis}
		\centering \text{d}_r^{p2}=\lvert\lvert \text{\textbf{h}}_p^2\text{M}_r^2 -\text{\textbf{t}}_p^2\rvert\rvert_{L_1/L_2}
	\end{equation}   
	where $\text{\textbf{h}}_p^2$ and $\text{\textbf{t}}_p^2$ denote corresponding element-wise projection in distance measurement space, $\text{M}_r^2$ is the transformation operation parameter matrix as $\text{M}_r^1$ via multiplying $M_{1}$, which is also a $\frac{k}{2}$*1 dimensional matrix. 
	
	Similarly, at different hierarchical level space, the corresponding measurement score can be obtained as Eq.\ref{hlevel}.
	
	\begin{equation}\label{hlevel}
		\centering \text{d}_r^{pk}=\lvert\lvert \text{\textbf{h}}_p^k\text{M}_r^k -\text{\textbf{t}}_p^k\rvert\rvert_{L_1/L_2}
	\end{equation}  
	
	Correspondingly, the semantic function of HIE at different deep level is defined as:
	\begin{equation}\label{deep_sem}
		\centering \text{d}_r^{sk}=\lvert\lvert \text{\textbf{h}}_s^k + \text{\textbf{r}}_s^k - \text{\textbf{t}}_s^k \rvert\rvert_{L_2}
	\end{equation}  
	
	And the total score function in deep level is defined as:
	\begin{equation}\label{deep}
		\centering \text{d}_r^k= \alpha \text{d}_r^{pk} + (1-\alpha) \text{d}_r^{sk}
	\end{equation}
	
	Finally, according to Eq.\ref{shallow} and Eq.\ref{deep}, the final score function of HIE is defined as follows.
	\begin{equation}\label{score}
		\centering f_r(\text{h},\text{t})=\sum_{k=1}^{N} \lambda_{k}\text{d}_r^{k}
	\end{equation}
	where $\lambda_{k}$ is the weights for the tradeoff between hierarchical level embedding scores, under the constrain $1=\sum \lambda_{k}$.
	
	\subsection{Loss Function}
	Following \cite{RotatE}, we use the negative sampling loss with self-adversarial training as the final loss and try to minimize it. The final loss function is illustrated as Eq.\ref{loss}.
	\begin{equation}\label{loss}
		L=-\text{log}\sigma(\gamma - f_r(\text{\textbf{h,t}}))-\sum_{j=1}^{n}p(\text{\textbf{h}}_j^{'},\text{\textbf{r}},\text{\textbf{t}}_j^{'})\text{log}\sigma(f_r(\text{\textbf{h}}_j^{'},\text{\textbf{t}}_j^{'})-\gamma)
	\end{equation}
	where $\gamma$  is a fixed margin, $\sigma$ denotes the sigmoid function, $(\text{h}_j^{'},\text{r},\text{t}_j^{'})$ is the $j$-th negative triplet, and $p(\text{h}_j^{'},\text{r},\text{t}_j^{'})$ is the negative triples sampling probability distribution, which is defined in Eq.\ref{neg}.
	\begin{equation}\label{neg}
		\centering
		p(\text{\textbf{h}}_i^{'},\text{\textbf{r}},\text{\textbf{t}}_i^{'}|\{\text{\textbf{h}}_j,\text{\textbf{r}}_j,\text{\textbf{t}}_j\})=\frac{\text{exp}\alpha f_r(\text{\textbf{h}}_i^{'},\text{\textbf{t}}_i^{'})}{\sum_j \text{exp}\alpha f_r(\text{\textbf{h}}_j^{'},\text{\textbf{t}}_j^{'})}
	\end{equation}
	where $\alpha$ is the temperature of sampling, and treated as the weight of negative samples.
	
	\subsection{Connection to other KGE models}
	Although some mathematical forms of our proposed HIE are similar to RotatE\cite{atuo}, in which the author uses modulus and phase information to learn entity and relation embeddings, the aims of the two models are different indeed. RotatE tries to model and infer all relation patterns, while HIE aims to obtain more expressive embeddings via extracting information from position distance space and semantic space. Moreover, RotatE can not deal with complex relations well due to the angle-representation. More specifically, RotatE is only an angle-specific representation of HIE by replacing the transformation matrix with a rotation angle in distance measurement space. Moreover, HIE utilizes hierarchical level information to imitate people's cognitive behavior to improve the expressive ability of the embedding process.     
	
	\section{Experiments}\label{myexp}
	
	
	
	
	\subsection{Datasets}
	\begin{table}[hbp]
		\caption{The statistics of datasets in the experiment. Rels denotes the number of relations. Ents is the number of entities. Train, Valid and Test denote the number of training, validation and test triples, respectively.}\label{tb2}
		\scalebox{0.9}{
			\begin{tabular}{cccccc}		
				\toprule
				\multirow{2}{*}{Datasets} & \multirow{2}{*}{Rels} &\multirow{2}{*}{Ents} & \multicolumn{3}{c}{Triples} \\
				\cline{4-6} 
				&   &   & Train & Valid & Test \\
				\toprule
				WN18 & 18 & 40,943 &141,442 & 5,000 & 5,000 \\
				WN18RR & 11 & 40,943 & 86,835 & 3,034 & 3,134\\
				FB15k-237 & 237 & 14,541 & 272,115 & 17,535 & 20,466\\
				YAGO3-10 & 37 & 123,182 & 1,079,040 & 5,000 & 5,000\\
				\bottomrule
			\end{tabular}
		}
	\end{table}
	
	Our proposed model HIE was evaluated on the following public benchmark datasets: WN18\cite{transE}, WN18RR\cite{conve}, FB15k-237\cite{convkb}, YAGO3-10\cite{RotatE}. All of them are extracted from the real-world knowledge graphs and contain massive entities and relations. The statistics of them are reported in Table.\ref{tb2}, and detailed information is listed as follows.
	
	\begin{enumerate}[(1)]
		\item WN18 is extracted from WordNet\cite{wordnet}, a lexical knowledge graph for English, consisting of 10 relations and 40,943 entities. Its entities represent the word sense, and the relations denote the lexical relationship between entities. Moreover, most the relation patterns are composed of $\mathbf{symmetry/antisymmetry}$ and $\mathbf{inversion}$.
		\item WN18RR is a subset of WN18, where all the inverse relations are deleted. It consists of 40,943 entities with 11 different relations. And the main relation patterns are $\mathbf{symmetry/antisymmetry}$ and $\mathbf{composition}$.
		\item FB15k-237 is a subset of FB15k with 14,541 entities and 237 relations. Besides, all the inverse relations are deleted. Thus, the main relation patterns are $\mathbf{composition}$ and $\mathbf{symmetry/antisymmetry}$.
		\item YAGO3-10 is extracted from YAGO3\cite{yaogo3}, which contains approximately 123,182 entities and 37 relations. It describes the relations among people and their attributes and has a minimum of 10 relations per. 
	\end{enumerate}
	\subsection{Evaluation Metrics}
	Link prediction is an essential task for knowledge graph completion, aiming to predict the missing entity according to a specific relation, i.e. (\textit{h, r}, ?) or (?,\textit{ r, t}). The real triplet can be obtained by ranking the results of score function $f_r$. Following the experimental settings in TransE\cite{transE}, for each given triplet $x_i$=(\textit{h,r,t})$\in$ T$_{test}$, we replace the head entity h or the tail entity t by any other entity in the knowledge graph to generate a set of corrupted triplets, i.e. $\widetilde{x}_i^s$=(\textit{h$^{cor}$,r,t})$\notin$T or  $\widetilde{x}_i^o$=(\textit{h,r,t$^{cor}$})$\notin$T respectively, where \textit{h$^{cor}$} and \textit{t$^{cor}$} denote a set of corresponding candidate entities. Then we check whether the proposed model HIE obtains a high score for the test triplet and a low score for corrupted triplets. The right ranks rank$_{i}^{o}$ and left ranks rank$_{i}^{s}$ of the $i-th$ test triplet $x_i$ are each associated with corrupting either head or tail entities corresponding to the score function Eq.\ref{score}, i.e. $f_r(\text{\textbf{h}},\text{\textbf{t}})$, and are defined as follows:
	
	\begin{equation}\label{rank}
		\centering
		\begin{split}
			rank_i^o&=1+\sum_{\widetilde{x}_i^o\notin \text{T}}I[\varphi(x_i) < \varphi(\widetilde{x}_i^o)]    \\
			rank_i^s&=1+\sum_{\widetilde{x}_i^s\notin \text{T}}I[\varphi(x_i) < \varphi(\widetilde{x}_i^s)]
		\end{split}
	\end{equation}
	where $I[P]$ denotes the indicator function. If the condition $P$ is true, the function returns 1 and returns 0 otherwise. In the experiment, the following three popular ranking metrics are utilized as our evaluation metrics: Mean rank (MR), Mean reciprocal rank (MRR) and Hits@k (k=1, 3, and 10 in this paper), which are defined in Eq.\ref{metrics}. MR and MRR denote the average ranks and the average inverse ranks for all test triplets, respectively. Hits@k is the percentage of ranks lower than or equal to k. In the above three metrics, higher MRR and Hits@k mean better performance, while the lower MR indicates better performance.
	\begin{equation}\label{metrics}
		\begin{split}
			&\text{MR: } \frac{1}{2G} \sum_{x_i \in \text{T}}(rank_i^o+rank_i^s) \\
			&\text{MRR: }\frac{1}{2G} \sum_{x_i \in  \text{T}}(\frac{1}{rank_i^o}+\frac{1}{rank_i^s}) \\
			&\text{Hits@k: } \frac{1}{2G} \sum_{x_i \in \text{T}}I[rank_i^o\le \text{k}]+ I[rank_i^s\le \text{k}]
		\end{split}
	\end{equation}
	where $G=\lvert \text{T}_{test} \rvert$ denotes the size of T$_{test}$.       
	
	\subsection{Implementation details}
	For this experiment, we implement our proposed model by Pytorch\cite{pytorch} with an adam optimizer\cite{adam}, and fine-tune the hyperparameters to obtain the optimal configuration via grid search approach. The hyperparameter settings are as follows: entity and relation embedding dimension $\in$ [250,500], batch size $\in$ [256,512], fixed margin $\gamma \in$ [3, 6, 9, 12, 18], self-adversarial sampling temperature $\alpha$ [0.5, 1.0], $\mathit{l}_1$-norm or $\mathit{l}_2$-norm, and test batch size $\in$ [8, 16], $\lambda_{1}$ and $\lambda_{2}$  $\in$ [ 0.2, 0.4, 0.6, 0.8]. 
	
	Moreover, we compare the proposed model HIE with three categories of state-of-the-art baselines: (1) translation-based models including TransE, TransR, TransD(unif), RotatE, TorusE and ComplEx. (2) semantic matching models, including DistMult. (3) deep neural network models including M-DCN, ConvE, ConvKB, KBGAN, R-GCN, HARN\cite{li2021learning} and InteractE. There are two criteria for the above algorithms to be selected, i.e., they achieve high performance within a certain range of comparison and provide the source code, or their architecture is easy for reproducibility.
	
	\subsection{Optimal hierarchical level}
\begin{figure}
		\centering
		\includegraphics[scale=0.35]{ 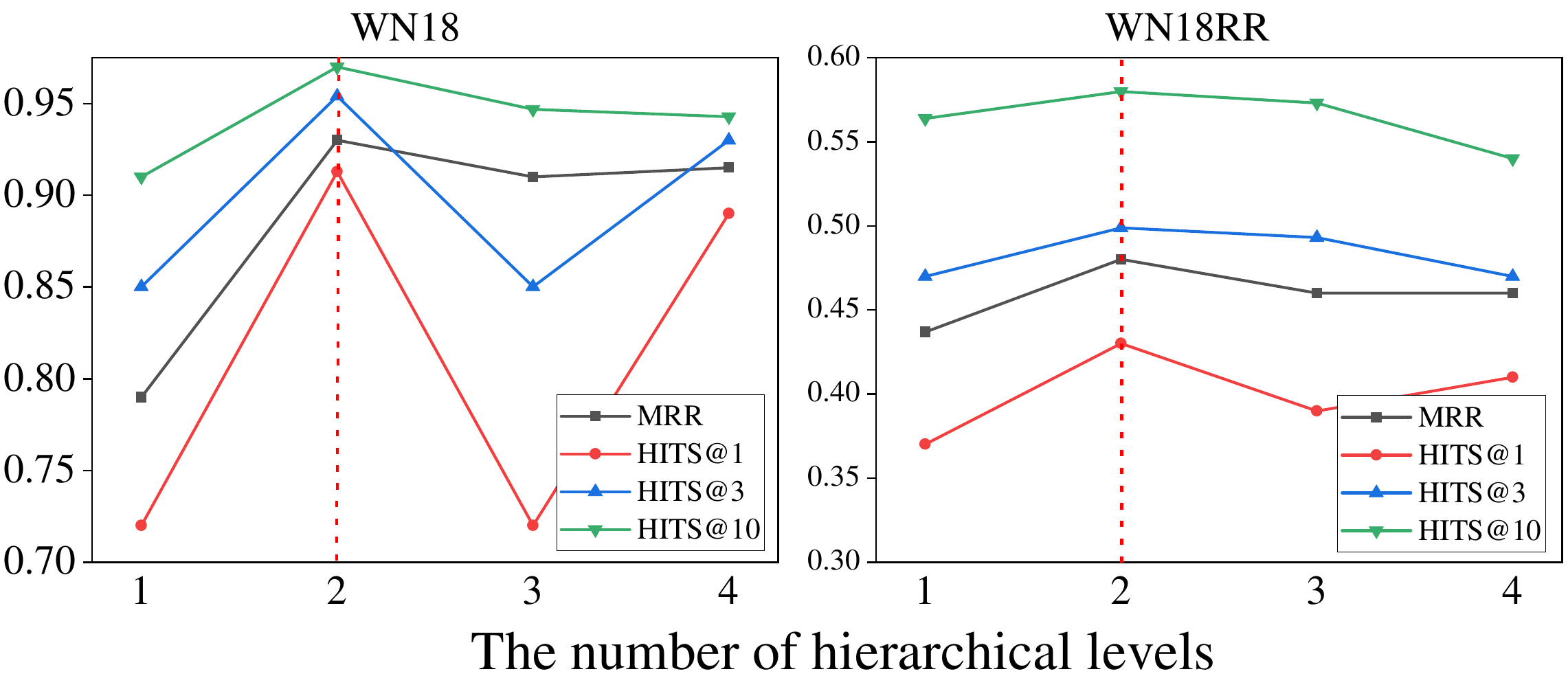}
		\caption{The results of different hierarchical level size on WN18 and WN18RR}
		\label{fig:wn18}
	\end{figure}

	Due to the limited computing resources, we need to determine the number of hierarchical levels at first. In this experiment, the link prediction results for hierarchical levels of HIE on WN18RR and WN18 are displayed in Fig.\ref{fig:wn18}. As the hierarchical level grows, the corresponding batch size and embedding dimension decrease, and the hierarchical weights are initialized equally. HIE with four different levels are implemented, as shown in Fig.\ref{fig:wn18}.
	
	As can be observed from Fig.\ref{fig:wn18}, all the metrics are optimal when two levels are set. This is mainly due to HIE can obtain structural information and semantic information from joint spaces efficiently. However, only one level for HIE is not adequate to capture enough information for modeling. Too many levels could also introduce noises, and may lead to poor performance for link prediction. Although the metrics in four level spaces are higher than that of three level spaces, the values of four evaluation metrics are still lower than that of two levels. In the next experiments, HIE with two levels are expressed as shallow level (level 1) and deep level (level 2). In conclusion, to balance the accuracy and resources utilization, HIE with two levels are chosen in the following experiments. Also, we can get the corresponding score function as defined in Eq.\ref{short}. 
	
	\begin{equation}\label{short}
		\centering f_r(\text{h},\text{t})=\lambda_{1}\text{d}_r^{1}+\lambda_{2}\text{d}_r^{2}
	\end{equation}

	\subsection{Link prediction}
	
	In this part, the link prediction results of HIE on four public datasets are summarized in Table.\ref{tblinkpre2} and Table.\ref{tblinkpre}. The best score is in $\textbf{bold}$ and the second-best score is \underline{underlined}. The evaluation results on WN18 and YGAO3-10 are reported in Table.\ref{tblinkpre2} and the results on WN18RR and FB15k-237 are shown in Table.\ref{tblinkpre}. Moreover, all the compared results are based on the optimal performance of the algorithm. 
	
	The observations from Table.\ref{tblinkpre2} can be summarized as follows:     
	\begin{itemize}
		\item Our proposed model can obtain significant overall evaluation scores from the empirical results compared with three kinds of KGE models on WN18 and YAGO3-10. Due to the massive inversion relation patterns existing in WN18 where the ratio reaches 94\%\cite{zhang2021knowledge}. To meet the inversion equation, the transformation matrix in HIE will be an identity matrix. That's to say, more information on distance measurement is omitted. Thus, the proposed HIE does not achieve best scores of MRR and Hits@1 on WN18.    
		\item On WN18, the proposed HIE outperforms the compared algorithms and achieves the best MR, Hits@3 and Hits@10. In terms of MRR and Hits@1, HIE is worse than that of several state-of-the-art models, while it obtains a 30\% higher Hits@1 than R-GCN\cite{rgcN} which uses more additional neighbor information. 
		\item On YAGO3-10, in terms of MRR, Hits@3 and Hits@10, HIE achieves the best score. As for the MR and Hits@1, HIE is slightly worse than the state-of-the-art RotatE and InteractE in the corresponding aspects. Notably, HIE obtains a 7\% higher MRR than that of neural network-based model M-DCN, 59\% higher MRR than that of semantic matching-based model DistMult, and 9\% higher than MRR that of translation based model RotatE. Strikingly, our proposed HIE achieves better performance scores on Hits@3 and Hits@10 than that of ConvE, RotatE and DistMult.        
	\end{itemize}
	
	\begin{table*}
		\caption{ Results of the link prediction on WN18 and YAGO3-10 datasets. Results [*] are taken from \cite{mdcn}. Results [$\clubsuit$] are from \cite{conve}. Other results are taken from the corresponding original papers.}\label{tblinkpre2}
		\scalebox{0.9}{
			\begin{tabular}{cccccccccccc}		
				\toprule
				\multicolumn{1}{c}{\multirow{3}{*}{Method}} & \multicolumn{5}{c}{WN18}    & \multicolumn{1}{c}{\multirow{3}{*}{}} & \multicolumn{5}{c}{YAGO3-10} \\ \cline{2-6} \cline{8-12} 
				\multicolumn{1}{c}{}                  & \multicolumn{1}{c}{\multirow{2}{*}{MRR$\uparrow$}} & \multicolumn{1}{c}{\multirow{2}{*}{MR$\downarrow$}} & \multicolumn{3}{c}{Hits@$\uparrow$}      & \multicolumn{1}{c}{}                  & \multicolumn{1}{c}{\multirow{2}{*}{MRR$\uparrow$}} & \multicolumn{1}{c}{\multirow{2}{*}{MR$\downarrow$}} & \multicolumn{3}{c}{Hits@$\uparrow$}      \\ \cline{4-6} \cline{10-12} 
				\multicolumn{1}{c}{}                  & \multicolumn{1}{c}{}                  & \multicolumn{1}{c}{}       & \multicolumn{1}{c}{1} & \multicolumn{1}{c}{3} & \multicolumn{1}{c}{10}         & \multicolumn{1}{c}{}                  & \multicolumn{1}{c}{}                  & \multicolumn{1}{c}{}        &\multicolumn{1}{c}{1}   & \multicolumn{1}{c}{3}       & \multicolumn{1}{c}{10}     \\  
				\toprule 
				TransE$^{[*]}$      & 0.454 & -   & 0.089 & 0.823 & 0.934 &  & 0.238 & -    & 0.212 & 0.361 & 0.447 \\
				TransR$^{[*]}$       & 0.605 & -   & 0.335 & 0.876 & 0.940 &  & 0.256 & -    & 0.223 & 0.356 & 0.478 \\
				TransD(unif)       & - & $\underline{229}$   & - & - & 0.925 &  & - & -    & - & - & - \\
				RotatE       & $\underline{0.949}$ & 309 & 0.944 & 0.952 & $\underline{0.959}$ &  & 0.495 & $\mathbf{1767}$ & 0.402 & 0.550 & 0.670 \\  
				DistMult$^{[\clubsuit]}$     & 0.822 & 902 & 0.728 & 0.914 & 0.936 &  & 0.340  & 5926 & 0.237 & 0.379 & 0.540 \\
				ComplEx$^{[\clubsuit]}$       & 0.941 & -   & 0.936 & 0.945 & 0.947 &  & 0.355 & 6351 & 0.258 & 0.399 & 0.547 \\
				M-DCN		 & $\mathbf{0.950}$ & -   & $\underline{0.946}$ & $\mathbf{0.954}$ & 0.958 &  & 0.505 & -    & 0.423 & $\underline{0.587}$ & 0.682 \\
				InteractE	 & -     & -   & -     & -     & -     &  & $\underline{0.541}$ & 2375 & $\mathbf{0.462}$ & -     & $\underline{0.687}$ \\
				ConvE        & 0.942 & 504   & $\mathbf{0.955}$ & $\underline{0.947}$ & 0.935 &  & 0.523 & 2792 & 0.448 & 0.564 & 0.658 \\  
				R-GCN        & 0.819 & -     &0.697  &0.929 &0.964 & &- &-&-&-&-\\
				\hline
				HIE          & 0.930  & $\mathbf{131}$  & 0.913  &  $\mathbf{0.954}$ & $\mathbf{0.970}$ & & $\mathbf{0.542}$ & $\underline{2042}$ & $\underline{0.452}$ & $\mathbf{0.593}$ & $\mathbf{0.695}$ \\
				\bottomrule
			\end{tabular}
		}
	\end{table*}
	
	From Table.\ref{tblinkpre}, we can conclude that:
	\begin{itemize}
		\item From the comparison of the experimental results on WN18RR and FB15k-237, compared with neural network-based models and semantic matching based models, we can observe that HIE achieves significant improvements for all the evaluation metrics. Also, it outperforms most translation-based models and obtains the best scores on MRR and Hits@3. That's to say. Our proposed HIE can capture more deep information than the previous neural network-based models and semantic information than translation-based and semantic matching based models.       
		\item On WN18RR, HIE achieves promising scores for each evaluation criteria. Compared with the neural network-based models, HIE obtains a 118\% higher MRR, 14\% higher Hits@10 than ConvKB, respectively. Strikingly, it obtains a 123\% higher MRR, 23.5\% higher Hits@10 than KBGAN. Also, HIE outperforms than DistMult and obtains a 18.2\% higher Hits@10. In terms of state-of-the-art translation-based algorithm RotatE, HIE is slightly worse in Hits@1. In contrast, HIE obtains better scores on MRR, MR, Hits@3 and Hits@10 than RotatE.    
		\item On FB15k-237, HIE is the best model in terms of MRR, Hits@1 and Hits@3. Besides, HIE obtains a 2.4\% higher MRR, 6\% higher Hits@1, and 1.9\% higher Hits@3 than that of RotatE. As for MR and Hits@10, HIE is slightly worse than RotatE.          
	\end{itemize}
	
	In a short summary, with the advantages of structural and semantic information, HIE can outperform the state-of-the-art algorithms on most of metrics.
	
	\begin{table*}
		\caption{ Results of the link prediction on WN18 and YAGO3-10 datasets. Results [$\heartsuit$] are taken from \cite{RotatE}.  Results [$\spadesuit$] are from the author's re-publication. Other results are taken from the corresponding original papers.}\label{tblinkpre}
		\scalebox{0.9}{
			\begin{tabular}{cccccccccccc}		
				\toprule
				\multicolumn{1}{c}{\multirow{3}{*}{Method}} & \multicolumn{5}{c}{WN18RR}    & \multicolumn{1}{c}{\multirow{3}{*}{}} & \multicolumn{5}{c}{FB15k-237} \\ \cline{2-6} \cline{8-12} 
				\multicolumn{1}{c}{}                  & \multicolumn{1}{c}{\multirow{2}{*}{MRR$\uparrow$}} & \multicolumn{1}{c}{\multirow{2}{*}{MR$\downarrow$}} & \multicolumn{3}{c}{Hits@$\uparrow$}      & \multicolumn{1}{c}{}                  & \multicolumn{1}{c}{\multirow{2}{*}{MRR$\uparrow$}} & \multicolumn{1}{c}{\multirow{2}{*}{MR$\downarrow$}} & \multicolumn{3}{c}{Hits@$\uparrow$}      \\ \cline{4-6} \cline{10-12} 
				\multicolumn{1}{c}{}                  & \multicolumn{1}{c}{}                  & \multicolumn{1}{c}{}       & \multicolumn{1}{c}{1} & \multicolumn{1}{c}{3} & \multicolumn{1}{c}{10}         & \multicolumn{1}{c}{}                  & \multicolumn{1}{c}{}                  & \multicolumn{1}{c}{}        &\multicolumn{1}{c}{1}   & \multicolumn{1}{c}{3}       & \multicolumn{1}{c}{10}     \\  
				\toprule 
				TransE$^{[\heartsuit]}$       & 0.226 & -    & - 	 & - 		& 0.501 &  & 0.294 & 357 & - & - & 0.465 \\
				MuRP   & 0.477 & -    & $\mathbf{0.438}$ 	 & 0.489 		& 0.555&  & 0.324 & - & 0.235 & 0.356 & 0.506 \\
				RotatE      				  & $\underline{0.476}$ & 3340 & 0.428 & $\underline{0.492}$    & $\underline{0.571}$ &  & $\underline{0.338}$ & $\mathbf{177}$ & $\underline{0.241}$ & $\underline{0.375}$ & $\mathbf{0.533}$ \\  
				DistMult     				  & 0.430 & 5110 & 0.390 & 0.440    & 0.490 &  & 0.241 & 254 & 0.155 & 0.263 & 0.419 \\
				TorusE 						  & 0.452 & -    & 0.422 & 0.464    & 0.512 &  & 0.305 & -   & 0.219 & 0.337 & 0.485 \\
				ConvKB$^{[\spadesuit]}$      				  & 0.220 &  $\mathbf{2741}$ & -     & -        & 0.508 &  & 0.302 & $\underline{196}$ & -     & -     & 0.483 \\
				KBGAN  						  & 0.215 & - &- & - &0.469   & & 0.277 &- & - & - &0.458 \\
				R-GCN        				  &- &-&-&-&- & & 0.249 & -     &0.151  &0.264 &0.417\\
				\hline
				HIE         				  & $\mathbf{0.480}$  & $\underline{2821}$  & $\underline{0.430}$  & $\mathbf{0.499}$ & $\mathbf{0.580}$ & & $\mathbf{0.346}$ & 215 & $\mathbf{0.255}$  &  $ \mathbf{0.380}$ & $\underline{0.523}$ \\
				\bottomrule
			\end{tabular}
		}
	\end{table*}
	
	
	\subsection{Complex Relations Modeling}
	In this part, the performance of HIE on several typical complex relations is reported. According to TransE\cite{transE}, relations can be classified into 1-to-1, N-to-1, 1-to-N and N-to-N. For each relation, it needs to compute the average number of head entities that are connected with specific tail entity $hco_r$ and the average number of tail entities that are connected with specific head entity $tcs_r$. Also, the complex relation patterns can be calculated as follows:
	
	\begin{equation}\label{compl_re}
		\centering
		\left\{\begin{array}{l} \text{hco}_r < \eta \ \text{and} \ \text{tcs}_r < \eta \Longrightarrow 1-\text{to}-1
			\\ \text{hco}_r < \eta \ \text{and} \ \text{tcs}_r \geqslant \eta \Longrightarrow \text{N}-\text{to}-1
			\\ \text{hco}_r \geqslant \eta \ \text{and} \ \text{tcs}_r < \eta \Longrightarrow 1-\text{to}-\text{N}
			\\ \text{hco}_r \geqslant \eta \ \text{and} \ \text{tcs}_r \geqslant \eta \Longrightarrow  \text{N}-\text{to}-\text{N}
		\end{array}\right.
	\end{equation}
	where $\eta$ = 1.5 is adopted the same as TransE\cite{transE}. 
	
\begin{figure}
		\centering
		\includegraphics[scale=0.25]{ 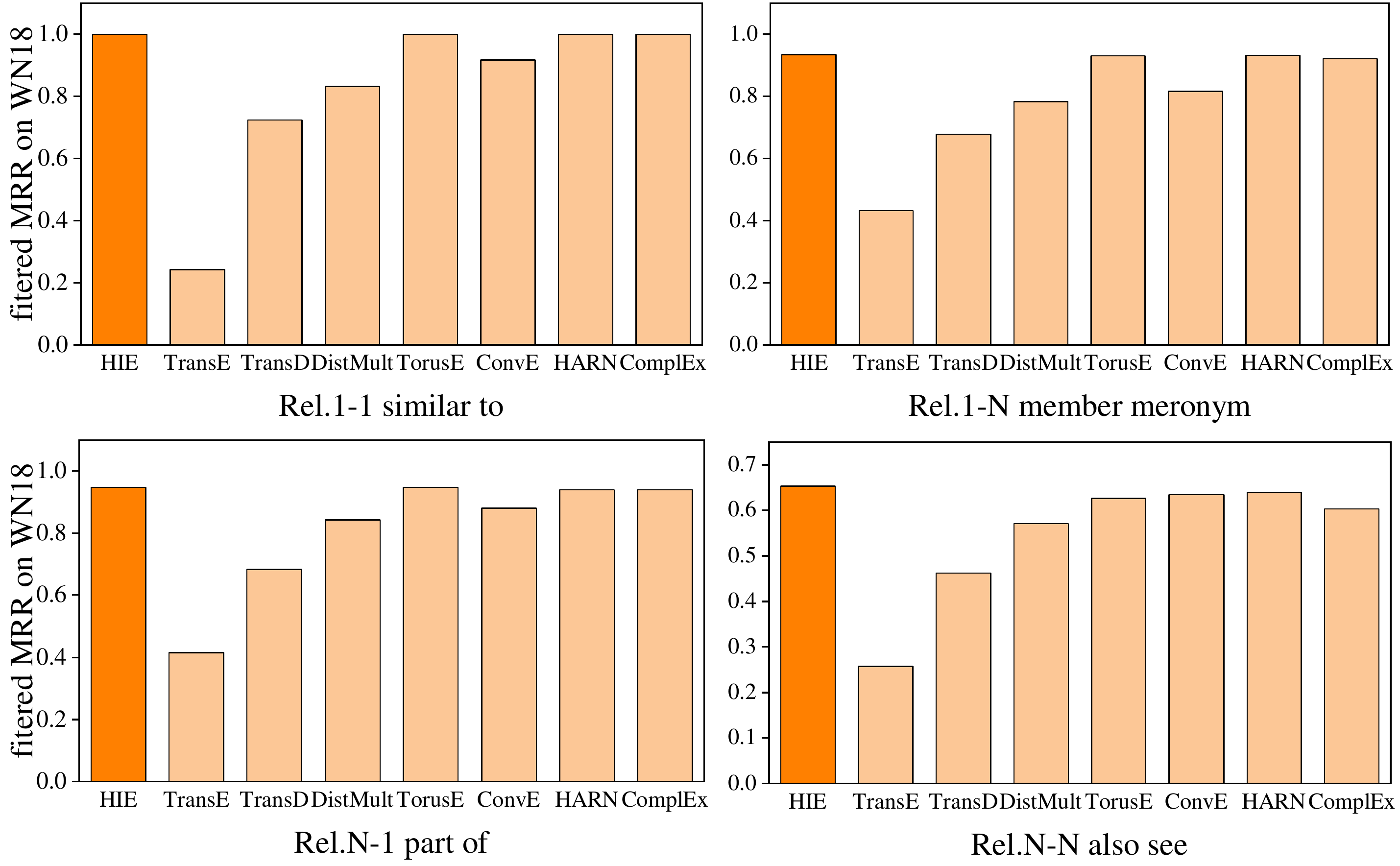}
		\caption{The results of part complex relations on WN18}
		\label{fig:5}
	\end{figure}

	Specifically, we extract four typical relationships in total corresponding to the specific relation patterns from WN18, i.e., $similar\_to$ (1-to-1), $member\_meronym$ (N-to-1), $part\_of$ (1-to-N), $also\_see$ (N-to-N). The results of MRR on the above four different relationships are displayed in Fig.\ref{fig:5}.
	
	Our proposed HIE is represented by orange, and the other baseline models are by light orange. Representative results can also be reported as follows:
	\begin{itemize}
		\item As can be seen from Fig.\ref{fig:5}, HIE can model the four kinds of relations on WN18. Compared with other models, HIE outperforms the compared algorithms with four relation categories.  
		\item In addition, HIE obtains a 0.758 higher MRR than TransE on $similar\_to$, a 19.3\% higher MRR than DistMult on $member\_meronym$, a 7.6\% higher MRR than ConvE on $part\_of$ and a 0.05 higher MRR than ComplEx on $also\_see$. 
		\item Translation-based and semantic matching-based models can not achieve promising results due to the lack of semantic information. However, neural network-based methods can capture more features of entities and relations via the convolution process. Thus, it achieves better performance than other baselines except for HIE. Take $also\_see$ as an example. HARN obtains a 0.383 higher MRR and 0.07 higher MRR than TransE and DistMult, respectively. However, with the advantage of deep level information and semantic information, HIE outperforms HARN and other neural network-based models as well.         
	\end{itemize}
	
	\subsection{Parameter Sensitivity Analysis}
	
	\subsubsection{study of $\lambda$}
\begin{figure}
		\centering
		\includegraphics[scale=0.3]{ 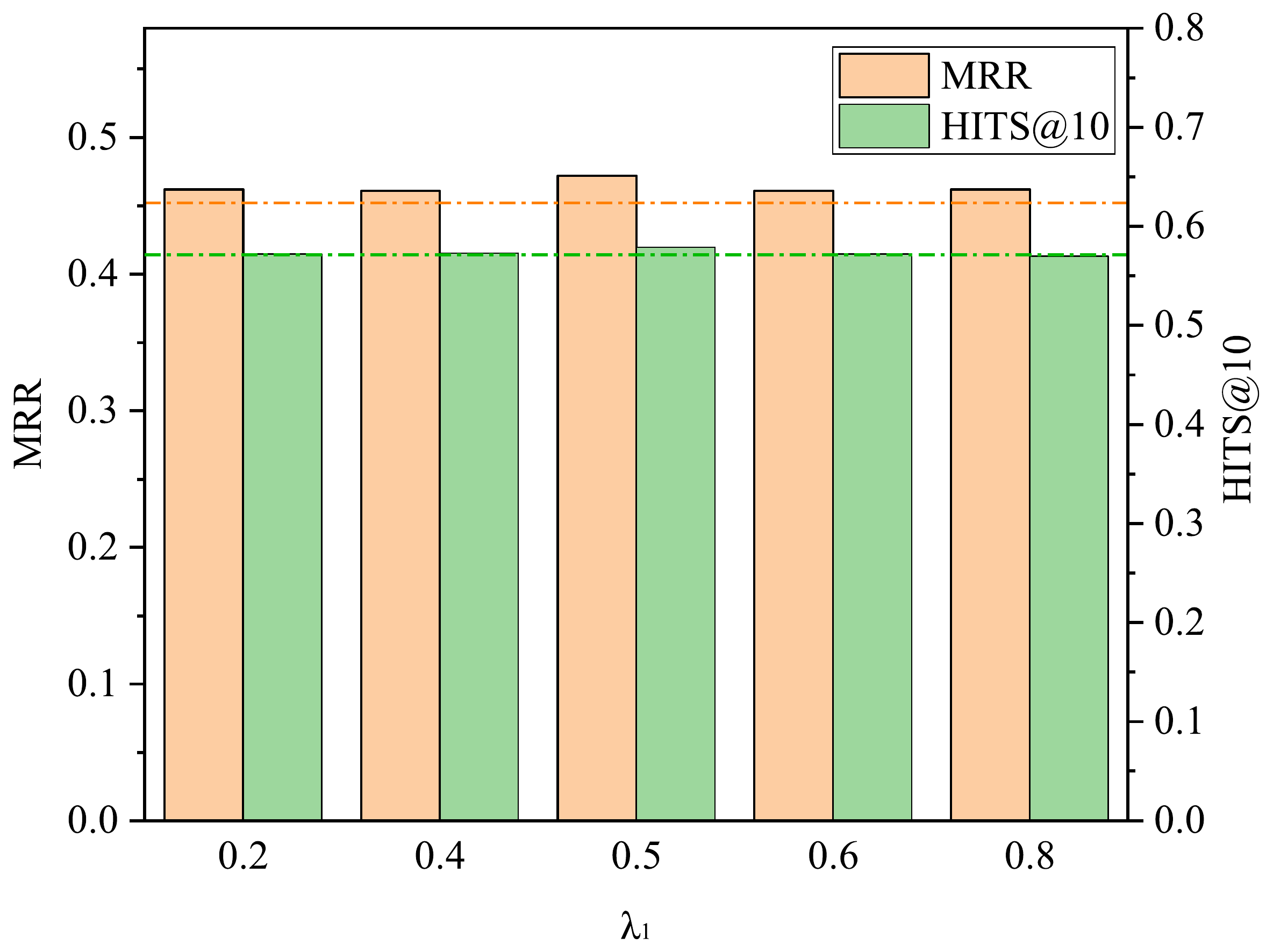}
		\caption{The results of different $\lambda_{1}$ on MRR and Hits@10 for WN18RR }
		\label{fig:lamda}
	\end{figure}

	The coefficient $\lambda_{1}$ and $\lambda_{2}$ are utilized to balance the shallow and deep level information and meet the condition $\lambda_{1}+\lambda_{2}=1$. Thus, we only investigated the effect of $\lambda_{1}$ for our proposed HIE on MRR and Hits@10 on WN18RR. The results are illustrated in Fig.\ref{fig:lamda}. The orange dotted line represents the MRR value of TorusE, and the green dotted line is the Hits@10 value of RotatE. We can observe that:
	\begin{itemize}
		\item When the $\lambda_{1}=0.5$, HIE achieves the best performance on MRR and Hits@10, and the values are 0.472 and 0.579, respectively. 
		\item Compared with TorusE, HIE obtains slightly higher MRR and Hits@10 when the $\lambda_{1}=0.2, 0.4, 0.6\ \text{and}\ 0.8$. In terms of $\lambda_{1}=0.5$, HIE outperforms TorusE by 5.5\%. In addition, HIE performs as good as RotatE when the $\lambda_{1}=0.2, 0.4\ \text{and}\ 0.6$, and slightly worse than RotatE. Also, HIE obtains a 1.4\% higher Hits@10 than that of RotatE. 
		\item The change of $\lambda_{1}$ performs a slight effect on MRR and Hits@10 of HIE. The reason is that HIE can capture more information from deep and shallow levels simultaneously and ensure the stability and accuracy.
	\end{itemize} 
	
	\subsubsection{study of batch size}
\begin{figure}
		\centering
		\includegraphics[scale=0.3]{ 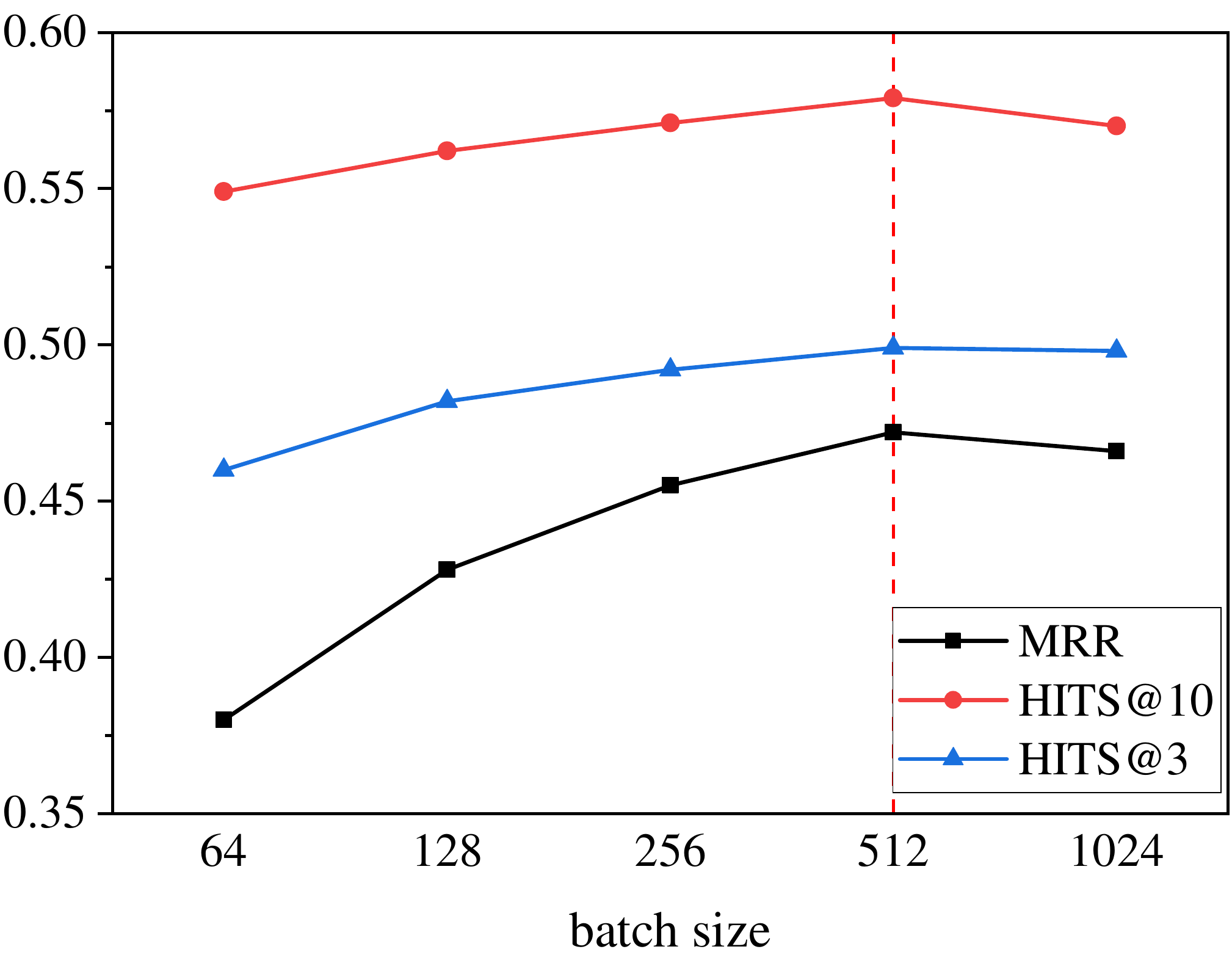}
		\caption{The results of different batch size on MRR for WN18RR}
		\label{fig:batchs}
	\end{figure}

	To explore the influence of batch size, we set different sizes as 64, 128, 256, 512 and 1024 with $\lambda_{1}=0.5$ on WN18RR. The results are reported in Fig.\ref{fig:batchs}. The red dotted line represents the best score on MRR, Hits@3 and Hits@10. We can see that the outstanding results can be achieved with batch size 512. Strikingly, as the batch size increases, HIE can obtain more information to improve itself. However, when the batch size is 1024, the MRR, Hits@3 and Hits@10 of HIE are worse than that of batch size 512,  which is mainly because the proposed HIE capture more and more information, including useful information as well as invalid information, even harmful information, to guide the optimization of the model. Simultaneously, larger information may limit the optimization capacity of HIE. Thus, batch size 512 is utilized in HIE to project entities and relations into low-dimensional space under the premise of measuring resources and accuracy. 
	
	\subsection{Ablation study}
	In this part, several ablation studies on WN18RR for HIE are carried out empirically. 

	\subsubsection{The impact of distance and semantic measurement}\label{sec462}
\begin{figure}
		\centering
		\includegraphics[scale=0.3]{ 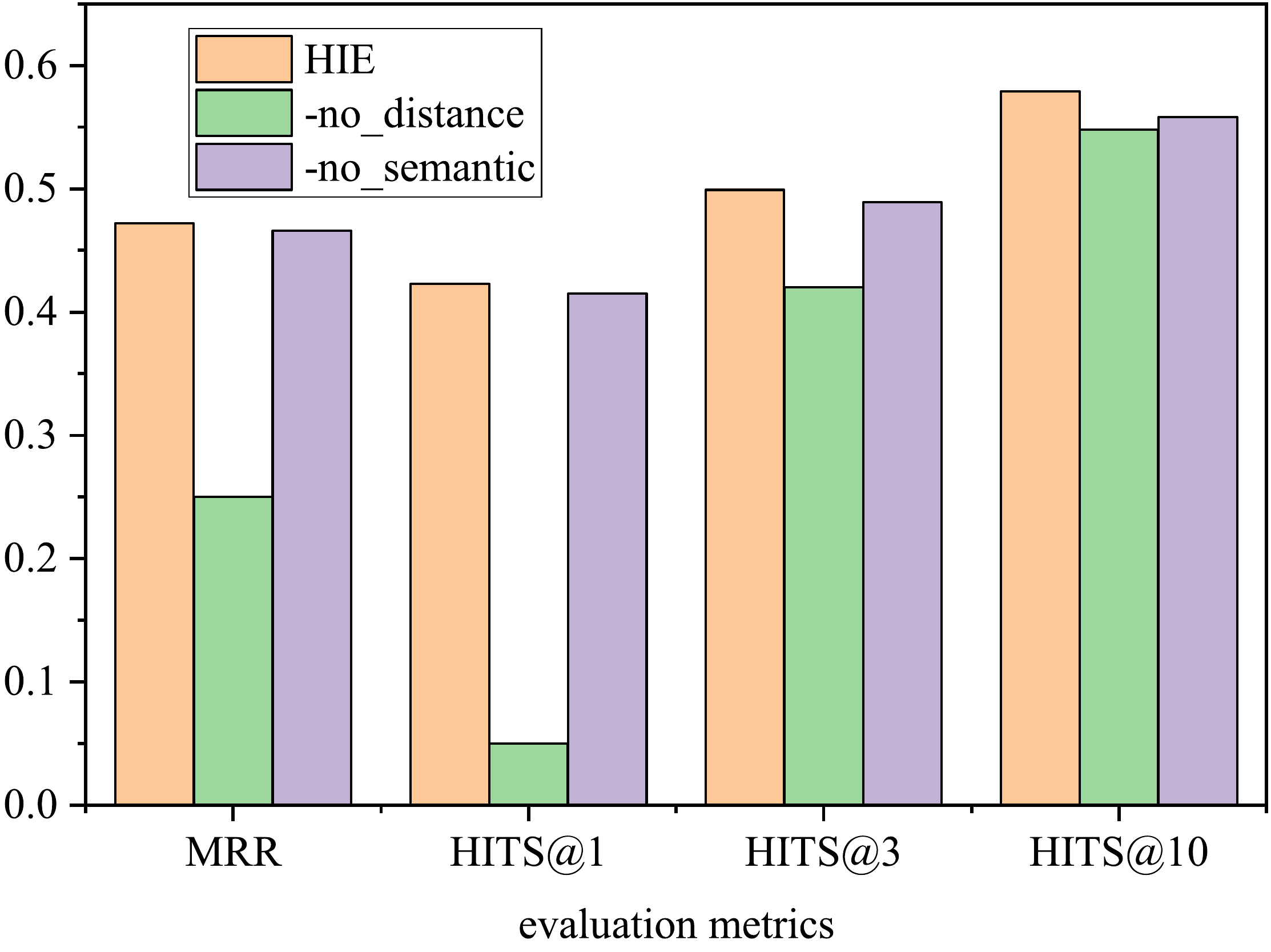}
		\caption{The ablation results of distance and semantic measurement}
		\label{fig:6}
	\end{figure}

	Fig.\ref{fig:6} shows the ablation study results on WN18RR of the impact of distance and semantic measurement when we only omit the distance measurement (-no\_distance) and omit the semantic measurement  (-no\_semantic) from HIE. As can be seen from Fig.\ref{fig:6}, three results can be reported as follows:
	\begin{itemize}
		\item Compared with the ablated model, the results demonstrate HIE achieves the best performance on all the evaluation metrics. Specifically, HIE obtains a 0.373 higher Hits@1 and a 0.21 higher Hits@10 than HIE-no\_distance and HIE-no\_semantic, respectively.
		\item In terms of MRR and Hits@1, distance measurement significantly impacts the performance of HIE. Also, HIE-no\_distance obtains a 16.2\% lower Hits@3 and a 2\% lower Hits@10 than HIE-no\_semantic.
		\item From the comparison of the empirical results, we can see that HIE-no\_semnatic or HIE-no\_distance could achieve promising scores on Hits@3 and Hits@10, which is mainly because of the use of deep level information. However, due to the lack of too much position information, HIE-no\_distance can not be well constrained in distance space. Thus, HIE-no\_distance has poor performance on Hits@1 and MRR. In comparison, HIE-no\_semantic is slightly worse than HIE.    
	\end{itemize}
	
	\subsubsection{The cross effects of both improvements}\label{sec463}
\begin{figure}
		\centering
		\includegraphics[scale=0.3]{ 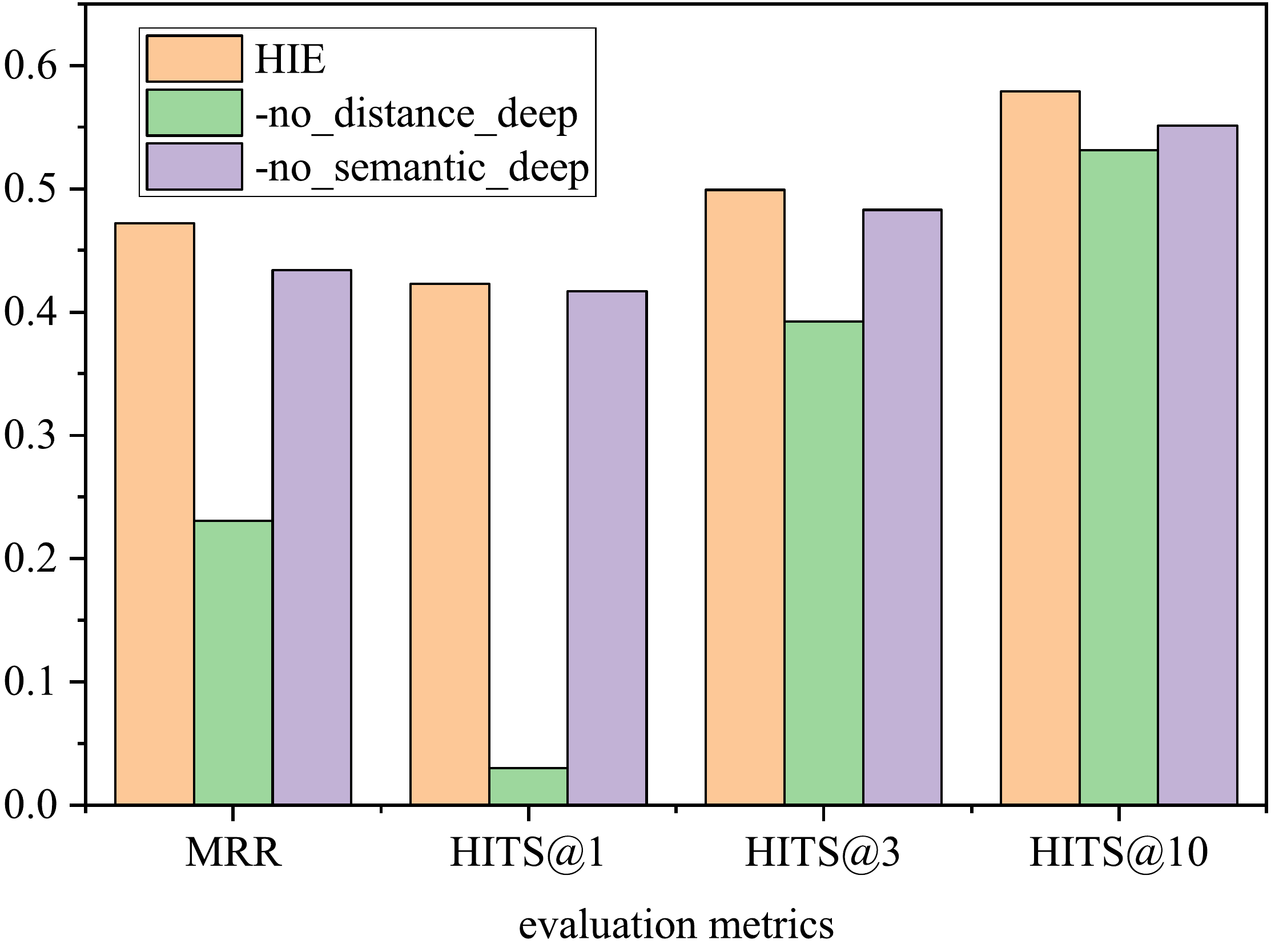}
		\caption{The ablation results of cross effects of both improvements}
		\label{fig:cross}
	\end{figure}

	Fig.\ref{fig:cross} shows the ablation study results on WN18RR of the cross effects of both improvements, i.e., distance measurement and its deep level information, and semantic measurement and its deep level information. As the results illustrated in Fig.\ref{fig:cross}, we can also see that:
	\begin{itemize}
		\item Distance measurement and deep level information play a crucial role in link prediction task. In contrast, semantic measurement can further mine the semantic information and improve the ability of the model.
		\item In terms of MRR, HIE achieves 9\% higher than HIE-no\_distance\_deep, while only 5.6\% higher than HIE-no\_distance. Also, without deep level information, HIE-no\_distance is even worse.
		\item Moreover, HIE obtains a 5.8\% higher Hits@10 than HIE-no\_semantic\_deep, while only a 3.7\% higher than that of HIE-no\_semantic, which can also demonstrate that deep level information could help improve the ability of HIE at a higher level.   
	\end{itemize}
	
	\section{Conclusion}\label{myconclusion}
	In this paper, we propose a novel KGE model to solve the link prediction task, namely, HIE. Each segmented entry of triplet (h, r, t) is projected into the distance and semantic space via the corresponding extractor. In addition, to capture more flexible structural information, a more general transformation operation matrix is utilized to obtain tail entity instead of translation-based operation. More importantly, we introduce hierarchical level information to learn knowledge representations from hierarchical spaces jointly. Our experiments on WN18, WN18RR, FB15k-237 and YAGO3-10 for the link prediction task show that HIE outperforms several state-of-the-art methods. Besides, the results of the complex relation modeling illustrate that HIE can better model the four typical relationships than existing neural network-based models. Specifically, the study of HIE further proves that hierarchical level information can significantly improve the performance of HIE, and both distance and semantic measurement could affect the expression ability of the model. In the future work, we will focus more on multi-task learning methods to capture more interactive information between entities and relations so as to improve the efficiency and performance of HIE.

	\bibliographystyle{elsarticle-num}
	\bibliography{my}

\end{document}